\documentclass[11pt,a4paper]{article}
\usepackage{naaclhlt2019,amsmath,graphicx}
\usepackage{times}
\usepackage{latexsym}
\usepackage{microtype}
\usepackage{multirow, multicol}
\usepackage{cite}
\usepackage{url}

\usepackage{tabularx}
\usepackage{booktabs}
\usepackage{multirow}
\usepackage{microtype}
\usepackage{subfigure}
\usepackage[utf8]{inputenc}

\newcolumntype{C}{>{\centering\arraybackslash}X}
\newcolumntype{L}{>{\raggedright\arraybackslash}X}

\newcolumntype{b}{X}
\newcolumntype{s}{>{\hsize=.2\hsize}X}

\newcolumntype{e}{>{\hsize=.4\hsize}X}

\newcolumntype{f}{>{\hsize=.5\hsize}X}

\usepackage{cite}

\usepackage[prependcaption,textsize=scriptsize]{todonotes}
\usepackage{xcolor}
\definecolor{mycolor}{HTML}{FF6600}

\aclfinalcopy 
\def\aclpaperid{202} 

\title{Pre-training on high-resource speech recognition \\ improves low-resource speech-to-text translation}

\author{
{\it Sameer Bansal$^{1}$, Herman Kamper$^{2}$, Karen Livescu$^{3}$, Adam Lopez$^{1}$, Sharon Goldwater$^{1}$}\\
$^{1}$School of Informatics, University of Edinburgh \\
$^{2}$E\&E Engineering, Stellenbosch University, South Africa \\
$^{3}$Toyota Technological Institute at Chicago, USA \\
{\tt\small \{sameer.bansal, sgwater, alopez\}@inf.ed.ac.uk, kamperh@sun.ac.za, klivescu@ttic.edu} \\
}

\date{}

\begin{document}
%
\maketitle
\begin{abstract}
We present a simple approach to improve direct speech-to-text translation (ST) when the source language is low-resource: we pre-train the model on a high-resource automatic speech recognition (ASR) task, and then fine-tune its parameters for ST. We demonstrate that our approach is effective by pre-training on 300 hours of English ASR data to improve Spanish-English ST from 10.8 to 20.2 BLEU when only 20 hours of Spanish-English ST training data 
are available. Through an ablation study, we find that the pre-trained encoder (acoustic model) accounts for most of the improvement, despite the fact that the shared language in these tasks is the target language text, not the source language audio. Applying this insight, we show that pre-training on ASR helps ST even when the ASR language differs from both source and target ST languages: pre-training on French ASR also improves Spanish-English ST.
Finally, we show that the approach improves 
performance on a true low-resource task: pre-training on a combination of English ASR and French ASR improves Mboshi-French ST, where only 4 hours of data are available, from 3.5 to 7.1 BLEU.

\end{abstract}
%
%
\section{Introduction}
\label{sec:intro}
Speech-to-text Translation (ST) has many potential applications for low-resource languages: for example in language documentation, where the source language is often unwritten or endangered~\citep{besacier2006towards,martin2015utterance,adams2016learning1,adams2016learning,anastasopoulos-chiang:2017:W17-01}; or in crisis relief, where emergency workers might need to respond to calls or requests in a foreign language~\citep{munro2010}. 
Traditional ST is a pipeline of automatic speech recognition (ASR) and machine translation (MT), and thus  requires transcribed source audio to train ASR and parallel text to train MT. These resources are often unavailable for low-resource languages, but for our potential applications, there may be some source language audio paired with target language text translations. In these scenarios, end-to-end ST is appealing.

Recently, \citet{weiss2017sequence} 
showed that end-to-end ST can be very effective, achieving an impressive BLEU score of 47.3 on Spanish-English ST. But this result required over 150 hours of translated audio for training, still a substantial resource requirement. By comparison, a similar system trained on only 20 hours of data for the same task achieved a BLEU score of 5.3 \citep{bansal2018interspeech}. Other low-resource systems have similarly low accuracies \citep{antonis+tied+naacl18,alexandre+audiobooks}.

To improve end-to-end ST in low-resource settings, we can try to leverage other data resources. For example, if we have transcribed audio in the source language, we can use multi-task learning to improve ST~\citep{antonis+tied+naacl18,weiss2017sequence,alexandre+audiobooks}. But source language transcriptions are unlikely to be available in our scenarios of interest.

Could we improve low-resource ST by leveraging data from a high-resource language? For ASR, training a single model on multiple languages can be effective for all of them \citep{toshniwal2018multilingual,deng2013recent}. For MT, \emph{transfer learning} \citep{Thrun1995IsLT} has been very effective:
pre-training a model for a high-resource language pair and transferring its parameters to a low-resource language pair when the target language is shared \citep{zoph2016transfer, johnson2016google}. Inspired by these successes, we show that low-resource ST can leverage transcribed audio in a high-resource target language, or even a different language altogether, simply by pre-training a model for the high-resource ASR task, and then transferring and fine-tuning some or all of the model's parameters for low-resource ST.

We first test our approach using Spanish as the source language and English as the target. After training an ASR system on 300 hours of English, fine-tuning on 20 hours of Spanish-English yields a BLEU score of 20.2, compared to only 10.8 for an ST model without ASR pre-training. Analyzing this result, we discover that the main benefit of pre-training arises from the transfer of the \emph{encoder} parameters, which model the input acoustic signal. In fact, this effect is so strong that we also obtain improvements by pre-training on a language that differs from either the source or target: pre-training on French and then fine-tuning on Spanish-English. We hypothesize that pre-training the encoder parameters, even on a different language, allows the model to better normalize over acoustic variability (such as speaker and channel differences), and conclude that this variability, rather than translation itself, is one of the main difficulties in low-resource ST. A final set of experiments confirm that ASR pre-training also helps on another language pair where the input is truly low-resource: Mboshi-French.

\section{Method}
\label{sec:method}
\begin{figure}[t]
  \centering
  \includegraphics[width=0.9\linewidth]{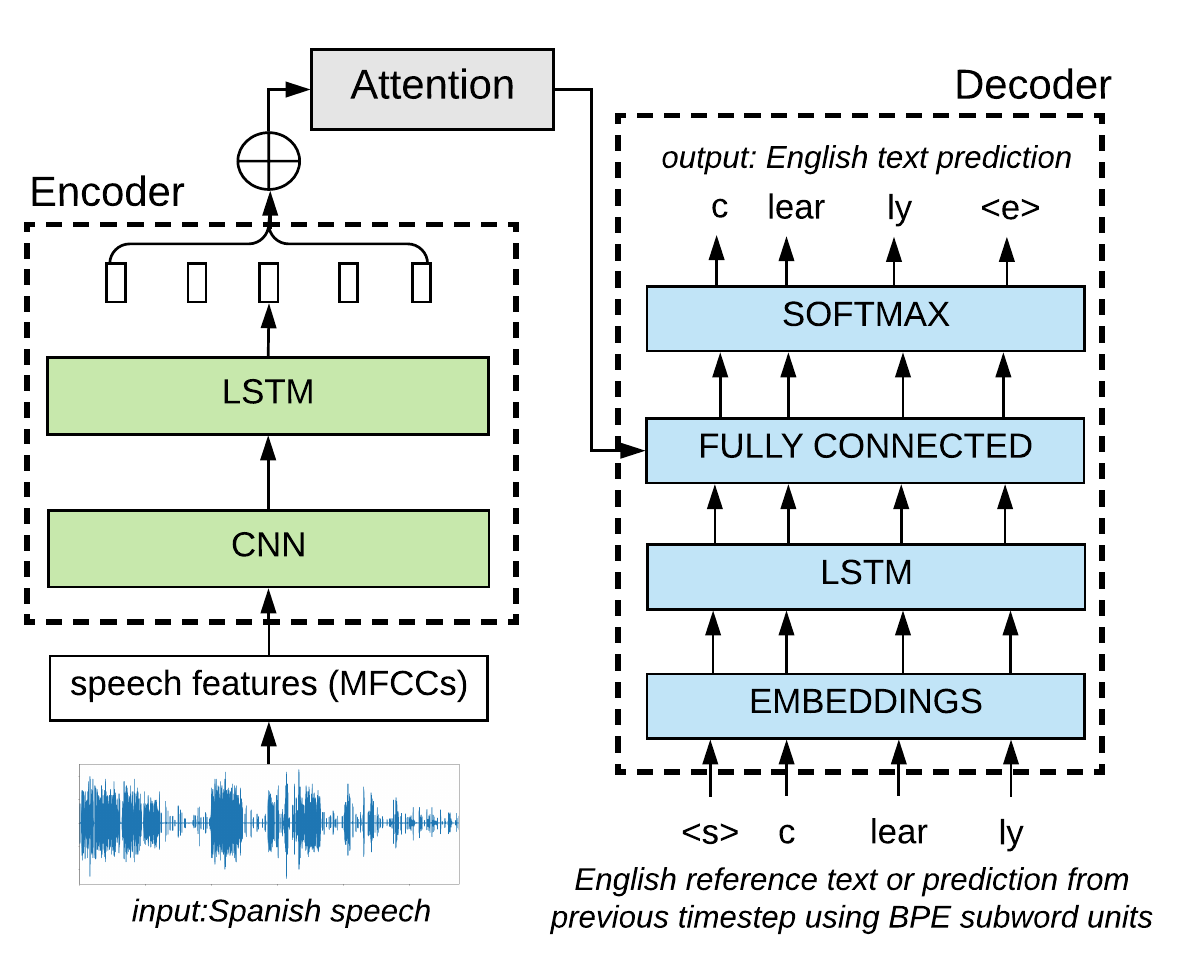}
  \caption{Encoder-decoder with attention model architecture for both ASR and ST. 
  The encoder input is the Spanish speech utterance {\em claro}, translated as {\em clearly},
  represented as BPE (subword) units.}
  \label{fig:enc_dec_arch}
\end{figure}

For both ASR and ST, we use an encoder-decoder model with attention adapted from \citet{weiss2017sequence}, \citet{alexandre+audiobooks} and \citet{bansal2018interspeech}, as shown in Figure~\ref{fig:enc_dec_arch}. We use the same model architecture for all our models, allowing us to conveniently transfer parameters between them. We also constrain the hyper-parameter search to fit a model into a single Titan X GPU, allowing us to maximize available compute resources.

We use a pre-trained English ASR model to initialize training of Spanish-English ST models, and a pre-trained French ASR model to initialize training of Mboshi-French ST models. In these configurations, the decoder shares the same vocabulary across the ASR and ST tasks. This is practical for settings where the target text language is high-resource with ASR data available.

In settings where both ST languages are low-resource, ASR data may only be available in a third language. To test whether transfer learning will help in this setting, we use a pre-trained French ASR model to train Spanish-English ST models; and English ASR for Mboshi-French models. In these cases, the ST languages are different from the ASR language, so we can only transfer the encoder parameters of the ASR model, since the dimensions of the decoder's output softmax layer are indexed by the vocabulary, which is not shared.\footnote{Using a shared vocabulary of characters or subwords is an interesting direction for future work, but not explored here.} Sharing only the speech encoder parameters is much easier, since the speech input can be preprocessed in the same manner for all languages. This form of transfer learning is more flexible, as there are no constraints on the ASR language used.

\section{Experimental Setup} 
\label{sec:experiments}

\subsection{Data sets} 
\label{sub:data sets}

\vspace{.1in}
\noindent {\bf English ASR.} We use the Switchboard Telephone speech corpus~\citep{LDC97S62}, which consists of around 300 hours of English speech and transcripts, split into 260k utterances. The development set consists of 5 hours that we removed from the training set, split into 4k utterances.

\vspace{.1in}
\noindent {\bf French ASR.} We use the French speech corpus from the GlobalPhone collection~\citep{schultz2002globalphone}, which consists of around 20 hours of high quality read speech and  transcripts, split into 9k utterances. The development set consists of 2 hours, split into 800 utterances.

\vspace{.1in}
\noindent {\bf Spanish-English ST.} We use the Fisher Spanish speech corpus~\citep{LDC2010S01}, which consists of 160 hours of telephone speech in a variety of Spanish dialects, split into 140K utterances. To simulate low-resource conditions, we construct smaller training corpora consisting of 50, 20, 10, 5, or 2.5 hours of data, selected at random from the full training data. The development and test sets each consist of around 4.5 hours of speech, split into 4K utterances. We do not use the corresponding Spanish transcripts; our target text consists of English translations that were collected through crowdsourcing \citep{post2013improved,LDC2014T23}.

\vspace{.1in}
\noindent {\bf Mboshi-French ST.} Mboshi is a Bantu language spoken in the Republic of Congo, with around 160,000 speakers.\footnote{\url{ethnologue.com/language/mdw}}
We use the Mboshi-French parallel corpus~\citep{mboshi+french+corpora}, which consists of around 4 hours of Mboshi speech, split into a training set of 5K utterances and a development set of 500 utterances. Since this corpus does not include a designated test set, we randomly sampled and removed 200 utterances from training to use as a development set, and use the designated development data as a test set.


\subsection{Preprocessing} 
\label{sub:preprocessing}

\vspace{.1in}
\noindent {\bf Speech.} We convert raw speech input to 13-dimensional MFCCs using Kaldi~\citep{povey+kaldi+asru+2011}.\footnote{In preliminary experiments, we did not find much difference between between MFCCs and more raw spectral representations like Mel filterbank features.}
We also perform speaker-level mean and variance normalization. 

\vspace{.1in}
\noindent {\bf Text.}
The target text of the Spanish-English data set
contains 1.5M word tokens and 17K word types. If we model text as sequences of words, our model cannot produce any of the unseen word types in the test data and is penalized for this, but it can be trained very quickly \citep{bansal2018interspeech}. If we instead model text as sequences of characters as in \citep{weiss2017sequence}, we would have 7M tokens and 100 types, resulting in a model that is open-vocabulary, but very slow to train \citep{bansal2018interspeech}. As an effective middle ground, we use byte pair encoding \citep[BPE;][]{sennrich-haddow-birch:2016:P16-12} to segment each word
into subwords, each of which is a character or a high-frequency sequence of characters---we use 1000 of these high-frequency sequences. Since the set of subwords includes the full set of characters, the model is still open vocabulary; but it results in a text with only 1.9M tokens and just over 1K types, which can be trained almost as fast as the word-level model.

The vocabulary for BPE depends on the frequency of character sequences, so it must be computed with respect to a specific corpus. For English, we use the full 160-hour Spanish-English ST target training text. For French, we use the Mboshi-French ST target training text. 



\subsection{Model architecture for ASR and ST} 
\label{sub:speech_to_text_model}

\vspace{.1in}
\noindent {\bf Speech encoder.} As shown schematically in Figure~\ref{fig:enc_dec_arch}, MFCC feature vectors are fed into a stack of two CNN layers, with 128 and 512 filters with a filter width of 9 frames each.
In each CNN layer we stride with a factor of 2 along time, apply a ReLU activation~\citep{nair2010rectified}, and apply 
batch normalization~\citep{ioffe+batchnorm+arxiv_2015}.
The output of the CNN layers is fed into a three-layer bi-directional LSTM~\citep{hochreiter+lstm}; each hidden layer has 512 dimensions.

\vspace{.1in}
\noindent {\bf Text decoder.} At each time step, the decoder chooses the most probable token from the output of a softmax layer produced by a fully-connected layer,
which in turn receives  the current state of a recurrent layer computed from previous time steps and an attention vector computed over the input. Attention is computed using the {\em global attentional model} with {\em general} score function and {\em input-feeding}, as described in~\citet{luong2015effective}. The predicted token is then fed into a 128-dimensional embedding layer followed by a three-layer LSTM to update the recurrent state; each hidden state has 256 dimensions. While training, we use the predicted token 20\% of the time as input to the next decoder step and the training token for the remaining 80\% of the time~\citep{williams+teacher_forcing}. At test time we use beam decoding with a beam size of 5 and length normalization~\citep{wu2016google+length+norm} with a weight of 0.6.

\vspace{.1in}
\noindent {\bf Training and implementation.} 
Parameters for the CNN and RNN layers are initialized using the scheme from~\citep{He+Normal}.
For the embedding and fully-connected layers, we use  Chainer's~\citep{chainer_learningsys2015} default initialition.

We regularize using dropout~\citep{srivastava+dropout}, with a ratio of $0.3$ over the embedding and LSTM layers~\citep{Gal2015Theoretically}, and a weight decay rate of $0.0001$. The parameters are optimized using Adam~\citep{kingma+adam+arxiv+2014}, with a starting alpha of 0.001.

Following some preliminary experimentation on our development set,
we add Gaussian noise with standard deviation of 0.25 to the MFCC features during training, and drop frames with a probability of 0.10. After 20 epochs, we corrupt the true decoder labels by sampling a random output label with a probability of 0.3. 

Our code is implemented in Chainer~\citep{chainer_learningsys2015} and we plan to make it freely available.



\subsection{Evaluation} 
\label{sub:evaluation}

\vspace{.1in}
\noindent {\bf Metrics.}
We report BLEU~\citep{papineni+bleu} for all our models.\footnote{We compute BLEU with {\tt multi-bleu.pl} from the Moses toolkit~\citep{koehn2007moses}.} In low-resource settings, BLEU scores tend to be low, difficult to interpret, and poorly correlated with model performance. This is because BLEU requires exact four-gram matches only, but low four-gram accuracy may obscure a high unigram accuracy and inexact translations that partially capture the semantics of an utterance, and these can still be very useful in situations like language documentation and crisis response. Therefore, we also report word-level unigram precision and recall, taking into account {\em stem}, {\em synonym}, and {\em paraphrase} matches. To compute these scores, we use METEOR~\citep{lavie+meteor} with default settings for English and French.\footnote{\url{cs.cmu.edu/~alavie/METEOR}} For example, METEOR assigns ``eat'' a recall of 1 against reference ``eat'' and a recall of 0.8 against reference ``feed'', which it considers a synonym match.

\vspace{.1in}
\noindent {\bf Naive baselines.}
We also include evaluation scores for a naive baseline model that predicts the {\em K} most frequent words of the training set as a bag of words for each test utterance. 
We set {\em K} to be the value at which precision/recall are most similar, which is always between 5 and 20 words. 
This provides an empirical lower bound on precision and recall, 
since we would expect any usable model to outperform a system that does not even depend on the input utterance. We do not compute BLEU for these baselines, since they do not predict sequences, only bags of words.



\section{ASR results}

\label{sub:asr_models}
Using the experimental setup of Section~\ref{sec:experiments}, we pre-trained ASR models in English and French, and report their word error rates (WER) on development data in Table~\ref{tab:wer_asr}.\footnote{We computed WER with the NIST {\tt sclite} script.}
We denote each ASR model by {\em L-Nh}, where {\em L} is a language code and {\em N} is the size of the training set in hours. For example, {\em en-300h} denotes an English ASR model trained on 300 hours of data.

Training ASR models for state-of-the-art performance requires substantial hyper-parameter tuning and long training times. 
Since our goal is simply to see whether pre-training is useful, 
we stopped pre-training our models after around 30 epochs (3 days) to focus on transfer experiments. As a consequence, our ASR results are far from state-of-the-art: current end-to-end Kaldi systems obtain 16\% WER on Switchboard {\em train-dev}, and 22.7\% WER on the French Globalphone dev set.\footnote{These WER results taken from respective Kaldi recipes on GitHub, and may not 
represent the very best results on these data sets.} We believe that better ASR pre-training may produce better ST results, but we leave this for future work.

\begin{table}
  \begin{center}
  \begin{tabularx}{0.9\linewidth}{CCCC}
    \toprule
      & {\bf en-100h} & {\bf en-300h} & {\bf fr-20h} \\
     \midrule
     {\bf WER} & 35.4 & 27.3 & 29.6 \\
  \bottomrule
  \end{tabularx}
  \end{center}
  \caption{Word Error Rate (WER, 
  in \%) for the ASR models used as pretraining, computed on Switchboard {\em train-dev} for English and Globalphone dev for French.
  }
  \label{tab:wer_asr}
\end{table}


\section{Spanish-English ST} 
\label{sec:spanish_english_results}

In the following, we denote an ST model by {\em S-T-Nh}, where {\em S} and {\em T} are source and target language codes, and {\em N} is the size of the training set in hours. For example, {\em sp-en-20h} denotes a Spanish-English ST model trained using 20 hours of data. We use the code {\em mb} for Mboshi and {\em fr} for French.

\subsection{Using English ASR to improve ST} 
\label{sub:spanish_english}

\begin{figure}[h]
\centering
\includegraphics[width=.45\textwidth]{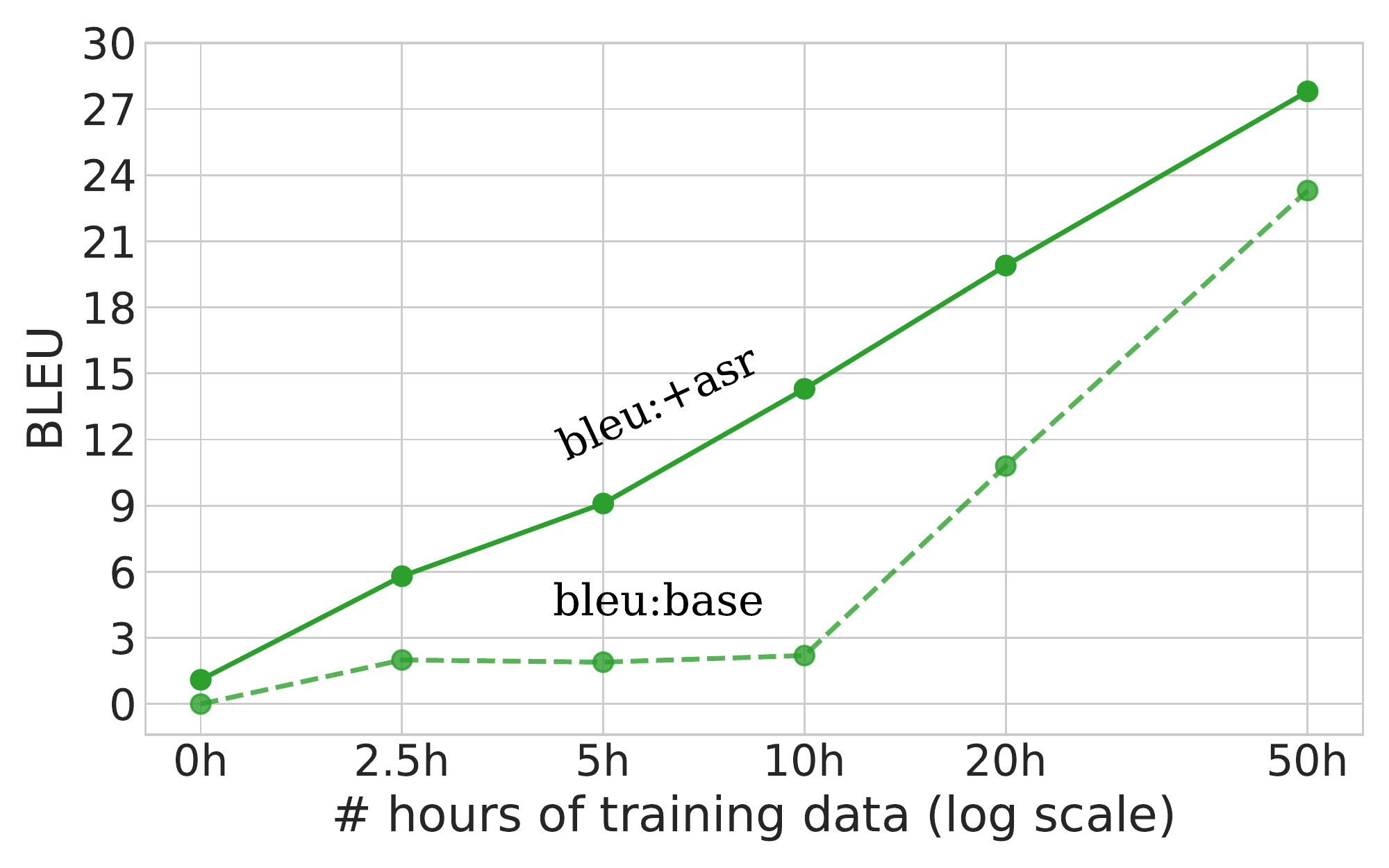}

\includegraphics[width=.45\textwidth]{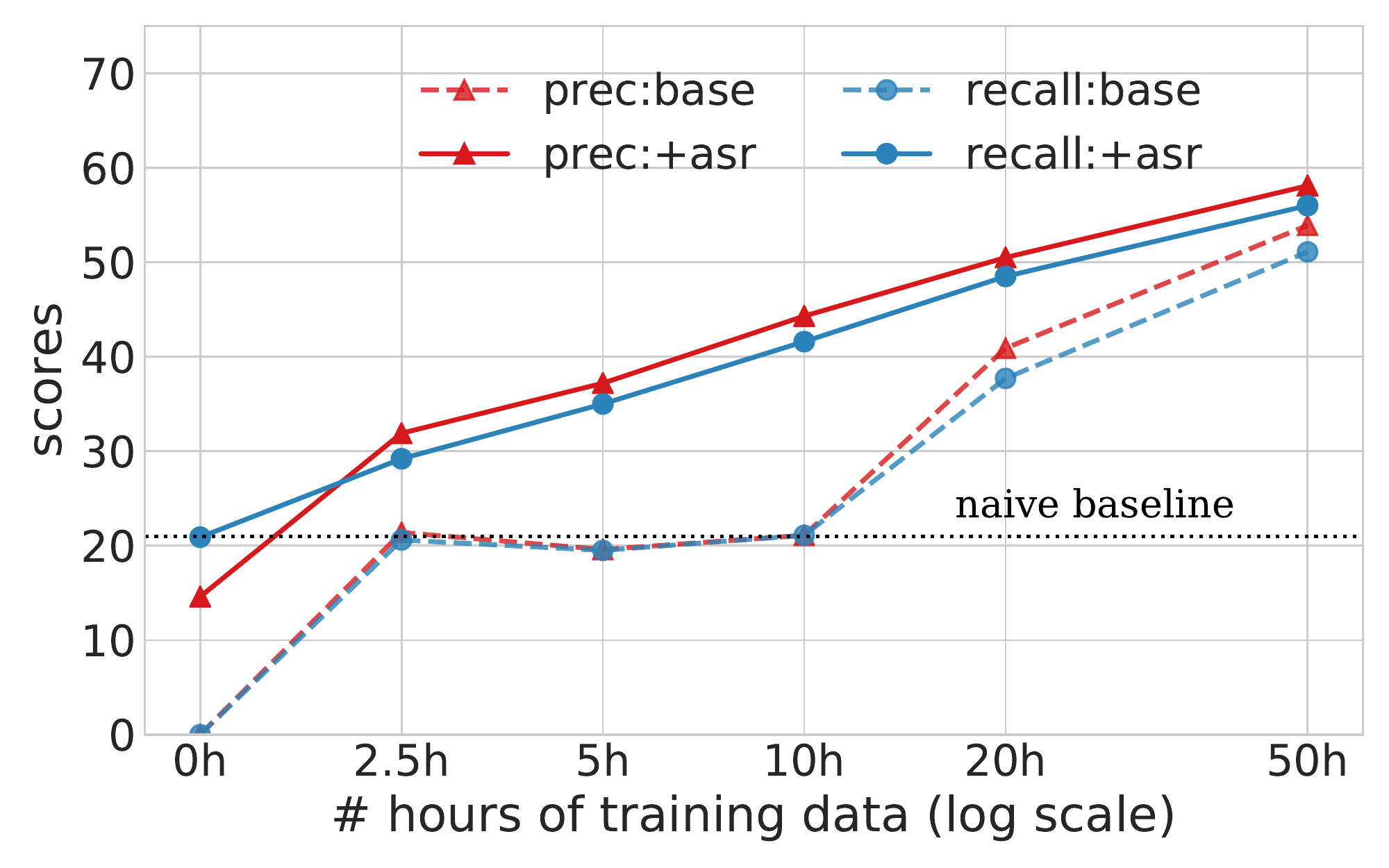}

\caption{(top) BLEU and (bottom) Unigram precision/recall for Spanish-English ST models computed on Fisher dev set. {\bf base} indicates no transfer learning; {\bf +asr} are models trained by fine-tuning {\em en-300h} model parameters. {\em naive baseline} indicates the score when we predict the 15 most frequent English words in the training set.}
\label{fig:sp_en_asr_bleu_prec_rec}
\end{figure}

Figure~\ref{fig:sp_en_asr_bleu_prec_rec} 
shows the BLEU and unigram precision/recall scores on the development set for baseline Spanish-English ST models and those trained after initializing with the {\em en-300h} model. Corresponding results on the test set (Table~\ref{tab:sp_en_bleu_test}) reveal very similar patterns. 
The remainder of our analysis is confined to the development set. The naive baseline, which predicts the 15 most frequent English words in the training set, achieves a precision/recall of around 20\%, setting a performance lower bound.

\vspace{.1in}
\noindent {\bf Low-resource: 20-50 hours of ST training data.}
Our baseline ST models substantially improve over previous results~\citep{bansal2018interspeech} using the same train/test splits, primarily due to  better regularization and modeling of subwords rather than words. Yet transfer learning still substantially improves 
over these strong baselines. For {\em sp-en-20h}, transfer learning improves dev set BLEU from 10.8 to 19.9, precision from 41\% to 51\%, and recall from 38\% to 49\%. For {\em sp-en-50h}, transfer learning improves BLEU from 23.3 to 27.8, precision from 54\% to 58\%, and recall from 51\% to 56\%.

\vspace{.1in}
\noindent {\bf Very low-resource: 10 hours or less of ST training data.}
Figure~\ref{fig:sp_en_asr_bleu_prec_rec} shows that without transfer learning, ST models trained on less than 10 hours of data struggle to learn, with precision/recall scores close to or below that of the naive baseline. But with transfer learning, we see gains in precision and recall of between 10 and 20 points.

\begin{table}[t]
  \begin{center}
  \begin{tabularx}{\linewidth}{rrrrrrr}
    \toprule
    {$N=$} & \multicolumn{1}{c}{0} & \multicolumn{1}{c}{2.5} & \multicolumn{1}{c}{5} & \multicolumn{1}{c}{10} & \multicolumn{1}{c}{20} & \multicolumn{1}{c}{50}  \\
     \midrule
     {\bf base} & 0 & 2.1 & 1.8 & 2.1 & 10.8 & 22.7 \\
     {\bf +asr} & 0.5 & 5.7 & 9.1 & 14.5 & 20.2 & 28.2 \\
  \bottomrule
  \end{tabularx}
  \end{center}
  \caption{BLEU scores for Spanish-English ST on the Fisher test set, using $N$ hours of training data. {\bf base}: no transfer learning. {\bf +asr}: using model parameters from English ASR (en-300h).}
  \label{tab:sp_en_bleu_test}
\end{table}

We also see that with transfer learning, a model trained on only 5 hours of ST data achieves a BLEU of 9.1, nearly as good as the 10.8 of a model trained on 20 hours of ST data without transfer learning. In other words, fine-tuning an English ASR model---which is relatively easy to obtain---produces similar results to training an ST model on four times as much data, which may be difficult to obtain.

We even find that in the very low-resource setting of just 2.5 hours of ST data, with transfer learning the model achieves a precision/recall of around 30\% and improves by more than 10 points over the naive baseline. In very low-resource scenarios with time constraints---such as in disaster relief---it is possible that even this level of performance may be useful, since it can be used to spot keywords in speech and can be trained in just three hours.

\vspace{.1in}
\noindent {\bf Sample translations.} Table~\ref{tab:sample_translations} shows example translations for models {\em sp-en-20h} and {\em sp-en-50h} with and without transfer learning using {\em en-300h}.

\begin{table}[t]
\begin{center}
  \footnotesize
  \begin{tabularx}{\linewidth}{l@{\hspace{1.5mm}}l}
    \toprule
    {\it Spanish} & super caliente pero muy bonito \\ 
    {\it English} &  super hot but very nice \\
     \noalign{\vskip 1.5mm}
    {\it 20h}  &  you support it \underline{but} it was \underline{very nice} \\
    {\it 20h+asr}  &  you can get alright \underline{but} it's \underline{very nice} \\
    {\it 50h} &  \underline{super} expensive \underline{but very nice} \\
    {\it 50h+asr}  & \underline{super hot but} it's \underline{very nice} \\

    \midrule
    {\it Spanish} & sí y usted hace mucho tiempo que que vive aquí \\ 
    {\it English} & yes and have you been living here a long time \\
     \noalign{\vskip 1.5mm}
    {\it 20h}  & \underline{yes} i've \underline{been} \underline{a long time} what did \underline{you} come \underline{here} \\
    {\it 20h+asr}  & \underline{yes and} \underline{you} \underline{have} \underline{a long time} that you \underline{live} \underline{here} \\
    {\it 50h} &  \underline{yes} \underline{you} are \underline{a long time} that you \underline{live} \underline{here} \\
    {\it 50h+asr}  & \underline{yes and have you been} \underline{here} \underline{long} \\
  \bottomrule
  \end{tabularx}
  \end{center}
  \caption{Example translations on selected sentences from the Fisher development set, with stem-level $n$-gram matches to the reference sentence underlined. {\bf 20h} and {\bf 50h} are Spanish-English models without pre-training; {\bf 20h+asr} and {\bf 50h+asr} are pre-trained on 300 hours of English ASR.}
  \label{tab:sample_translations}
\end{table}

\begin{figure}[h]
\centering
\subfigure[{\bf 50h:baseline}]{
\includegraphics[width=.40\textwidth]{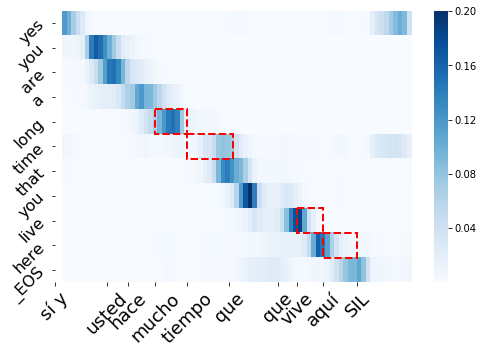}
}

\subfigure[{\bf 50h:asr}]{
\includegraphics[width=.40\textwidth]{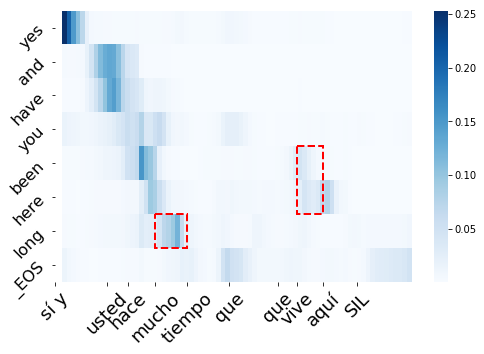}
}
\caption{
Attention plots for the final example in Table~\ref{tab:sample_translations}, using 50h models with and without pre-training. The $x$-axis shows the reference Spanish word positions in the input; 
the $y$-axis shows the predicted English subwords. In the reference, {\it mucho tiempo} is translated to {\it long time}, and {\it vive aquí} to {\it living here}, but their order is reversed, and this is reflected in (b). 
}
\label{fig:attn_plot_50h}
\end{figure}

Figure~\ref{fig:attn_plot_50h} 
shows the attention weights for the last sample utterance in Table~\ref{tab:sample_translations}. For this utterance, the Spanish and English text have a different word order: {\it mucho tiempo} occurs in the middle of the speech utterance, and its translation, {\it long time}, is at the end of the English reference. Similarly, {\it vive aquí} occurs at the end of the speech utterance, while the translation, {\it living here}, is in the middle of the English reference. The baseline {\em sp-en-50h} model translates the words correctly but doesn't get the English word order right. With transfer learning, the model produces a shorter but still accurate translation in the correct word order.

\subsection{Analysis} 
\label{sub:analysis}

To understand the source of these improvements, we carried out a set of ablation experiments. For most of these experiments, we focus on Spanish-English ST with 20 hours of training data, with and without transfer learning.

\vspace{.1in}
\noindent {\bf Transfer learning with selected parameters.} In our first set of experiments, we transferred all parameters of the {\em en-300h} model, including the speech encoder CNN and LSTM; the text decoder embedding, LSTM and output layer parameters; and attention parameters.
To see which set of parameters has the most impact, we train the {\em sp-en-20h} model by transferring only selected parameters from {\em en-300h}, and randomly initializing the rest. 

\begin{figure}[t]
  \centering
  \includegraphics[width=0.9\linewidth]{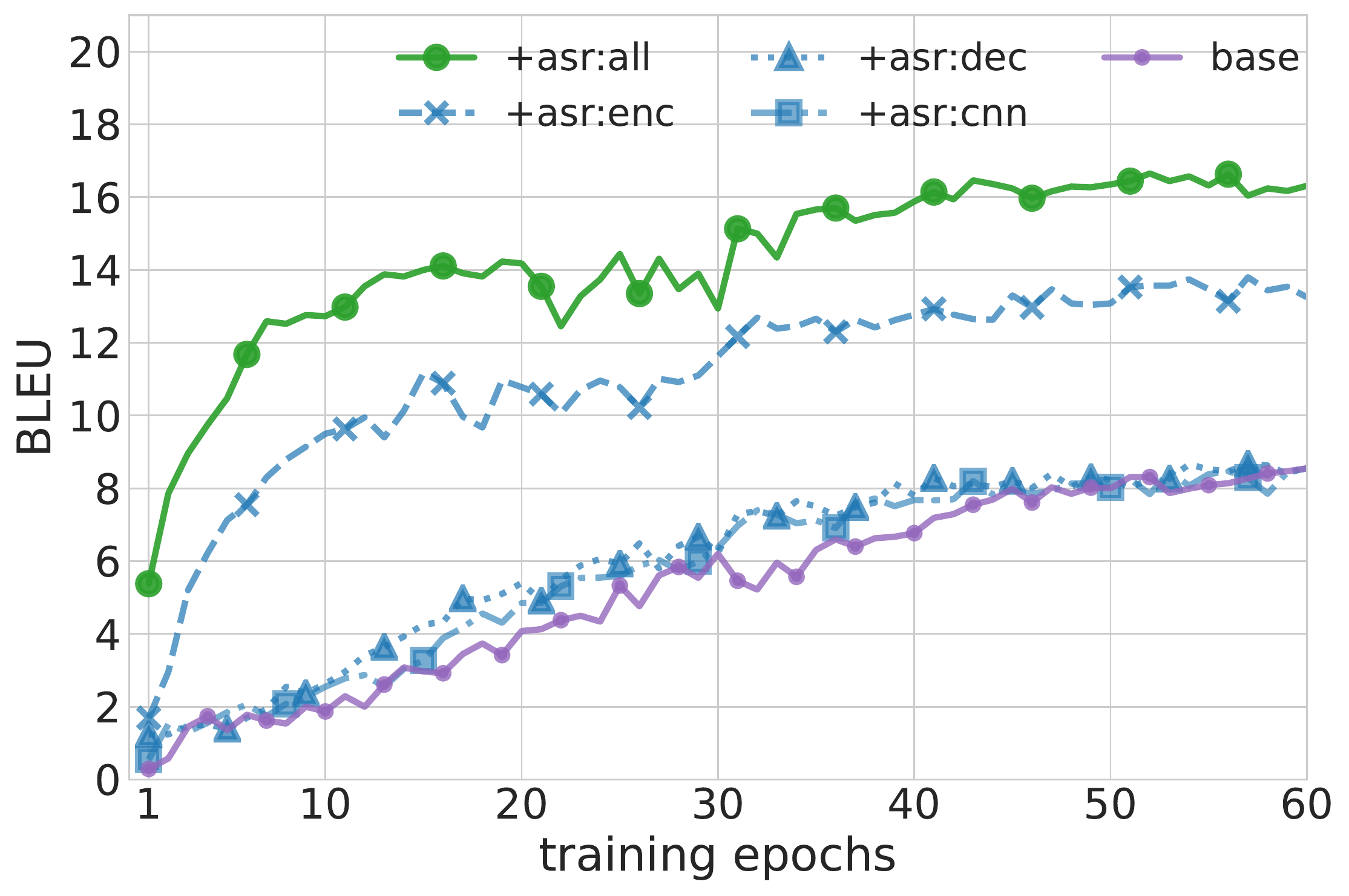}
  \caption{Fisher development set training curves (reported using BLEU) for {\em sp-en-20h} using selected parameters from {\em en-300h}: none ({\bf base}); encoder CNN only ({\bf +asr:cnn}); encoder CNN and LSTM only ({\bf +asr:enc}); decoder only ({\bf +asr:dec}); and all: encoder, attention, and decoder ({\bf +asr:all}). These scores do not use beam search and are therefore lower than the best scores reported in Figure~\ref{fig:sp_en_asr_bleu_prec_rec}.}
  \label{fig:bleu_asr_params}
\end{figure}

The results (Figure~\ref{fig:bleu_asr_params}) show that transferring all parameters is most effective. But they also show that the speech encoder parameters account for most of the gains. We hypothesize that the encoder learns transferable low-level acoustic features that normalize across variability like speaker and channel differences, and that much of this learning is language-independent. This hypothesis is supported by other work showing the benefits of cross-lingual and multilingual training for speech technology in low-resource target languages~\citep{carlin+2011+rapideval,jansen2010towardsST,deng2013recent,vu2012investigation,thomas2012multilingual,cui2015multilingual,alumae2016improved,yuan2016learning,renshaw2015comparison}. 
Indeed, there is evidence that speech features trained on multiple languages transfer better than those trained on the same amount of data from a single language~\citep{enno2018interspeech}. 

By contrast, transferring only decoder parameters does not improve accuracy.
Since decoder parameters help when used in tandem with encoder parameters, we suspect that the dependency in parameter training order might explain this: the transferred decoder parameters have been trained to expect particular input representations from the encoder, so transferring only the decoder parameters without the encoder might not be useful. 

Figure~\ref{fig:bleu_asr_params} also suggests that models make strong gains early on in the training when using transfer learning. The {\em sp-en-20h} model initialized with all model parameters ({\em +asr:all}) from {\em en-300h} reaches a higher BLEU score after just 5 epochs (2 hours) of training than the model without transfer learning trained for
60 epochs/20 hours. This again can be useful in disaster-recovery scenarios, where the time to deploy a working system must be minimized.

\begin{figure}[t]
  \centering
  \includegraphics[width=0.9\linewidth]{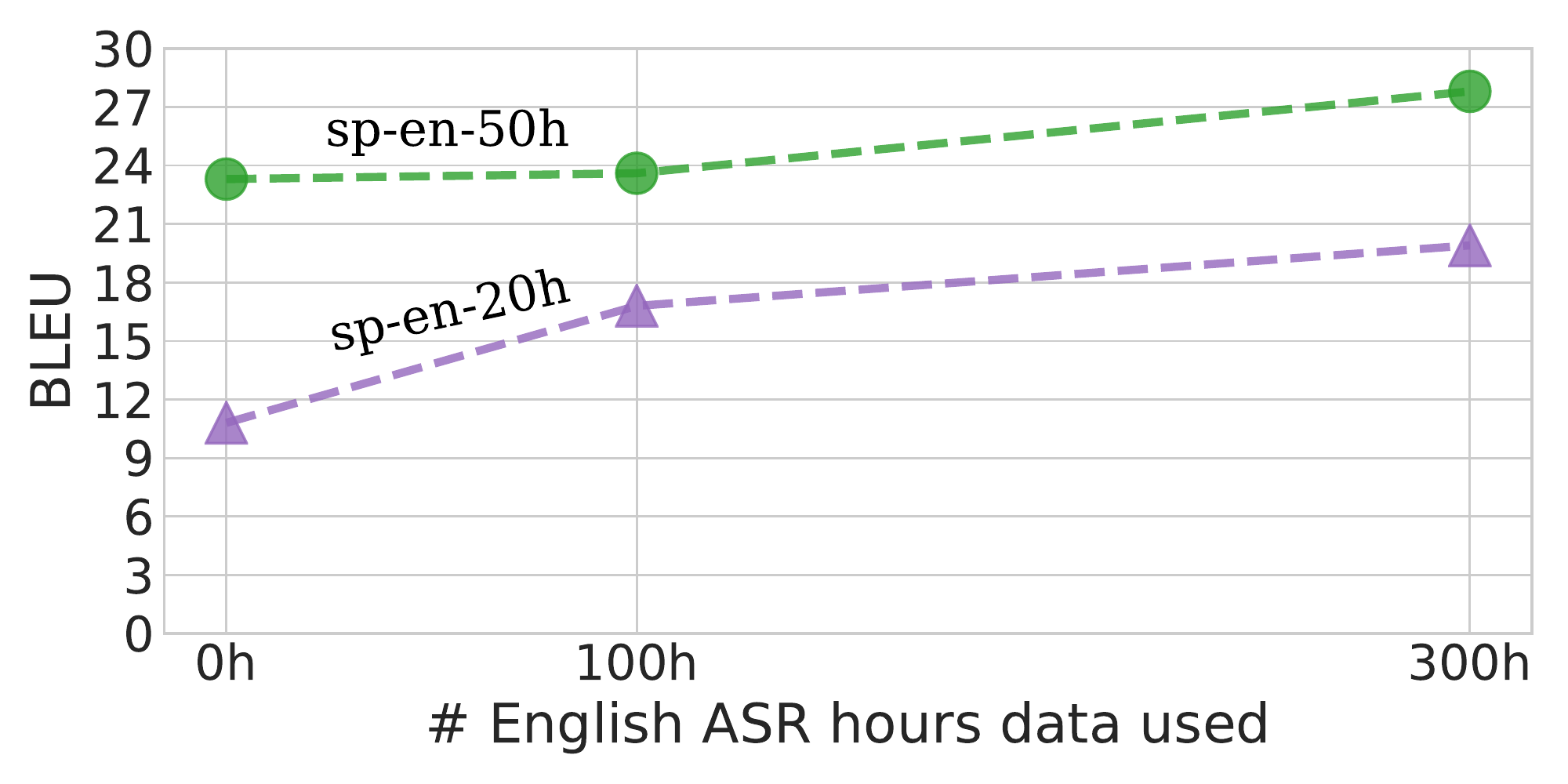}
  \caption{Spanish-to-English {\bf BLEU} scores on Fisher dev set, with 0h (no transfer learning), 100h and 300h of English ASR data used.}
  \label{fig:bleu_asr_data}
\end{figure}

\vspace{.1in}
\noindent {\bf Amount of ASR data required.} Figure~\ref{fig:bleu_asr_data} shows the impact of increasing the amount of English ASR data used on Spanish-English ST performance for two models: {\em sp-en-20h} and {\em sp-en-50h}. 

For {\em sp-en-20h}, we see that using {\em en-100h} improves performance by almost 6 BLEU points. By using more English ASR training data ({\em en-300h}) model, the BLEU score increases by almost 9 points.
However, for {\em sp-en-50h}, we only see improvements when using {\em en-300h}. This implies that transfer learning is most useful in true low-resource scenarios, when only a few tens of hours of training data are available for ST.
As the amount of ST training data increases, the benefits of transfer learning tail off, although it's possible that using even more monolingual data, or improving the training at the ASR step, could extend the benefits to larger ST data sets.

\vspace{.1in}
\noindent {\bf Impact of code-switching.}
We also tried using the {\em en-300h} ASR model without any fine-tuning to translate Spanish audio to English text. This model achieved a BLEU score of 1.1, with a precision of 15 and recall of 21. The non-zero BLEU score indicates that the model is matching \emph{some} 4-grams in the reference. This seems to be due to code-switching in the Fisher-Spanish speech data set. Looking at the dev set utterances, we find several examples where the Spanish transcriptions match the English translations, indicating that the speaker switched into  English. For example, there is an utterance whose Spanish transcription and English translation are both ``right yeah'', and this English expression is indeed present in the source audio. The English ASR model correctly translates this utterance, which is unsurprising since the phrase ``right yeah'' occurs nearly 500 times in Switchboard.

Overall, we find that nearly 500 of the 4,000 development set utterances (14\%) likely contain code-switching, since the Spanish transcription and English translations share more than half of their tokens. This suggests that transfer learning from English ASR models might help more than from other languages. To isolate this effect from transfer learning of language-independent speech features, we carried out a further experiment.


\subsection{Using French ASR to improve Spanish-English ST} 
\label{sub:using_french_asr_to_improve_spanish_english_st}
In this experiment, we pre-train using French ASR data for a Spanish-English translation task. Here, we can only transfer the speech encoder parameters, and there should be little if any benefit due to code-switching.

Because our French data set (20 hours) is much smaller than our English one
(300 hours), for a fair comparison we used a 20 hour subset of the English data for pre-training in this experiment. For both the English and French models, we transferred only the encoder parameters.

Table~\ref{tab:bleu_sp_en_cross} shows that both the English and French 20-hour pre-trained models improve performance on Spanish-English ST. The English model works slightly better, as would be predicted given our discussion of code-switching, but the French model is also useful, improving BLEU from 10.8 to 12.5. This result strengthens the claim that ASR pre-training on a completely distinct third language can help low-resource ST. Presumably benefits would be much greater if we used a larger ASR data set, as we did with English above.

\begin{table}
  \begin{center}
  \begin{tabularx}{0.95\linewidth}{rccc}
    \toprule
     & \multicolumn{1}{c}{\bf baseline} & \multicolumn{1}{c}{\bf +fr-20h} & \multicolumn{1}{c}{\bf +en-20h}\\
     \midrule
     sp-en-20h & 10.8 &  12.5 & 13.2 \\
    \bottomrule
  \end{tabularx}
  \end{center}
  \caption{
  Fisher dev set BLEU scores for {\em sp-en-20h}. {\bf baseline}: model without transfer learning. 
  Last two columns: Using encoder parameters from French ASR ({\bf +fr-20h}), and English ASR ({\bf +en-20h}).
  }
  \label{tab:bleu_sp_en_cross}
\end{table}

\section{Mboshi-French ST} 
\label{sec:mboshi_french}

Our final set of experiments test our transfer method on ST for the low-resource language Mboshi, where we have only 4 hours of ST training data: Mboshi speech input paired with French text output.

Table~\ref{tab:bleu_mb_fr_st} shows the ST model scores for Mboshi-French with and without using transfer learning. The first two rows {\em fr-top-8w}, {\em fr-top-10w}, show precision and recall scores for the {\em naive baselines} where we predict the top 8 or 10 
most frequent French words in the Mboshi-French training set. These show that a precision/recall in the low 20s is easy to achieve, although with no n-gram matches (0 BLEU). The pre-trained ASR models by themselves (next two lines) are much worse.

\begin{table}[t]
  \begin{center}
  \begin{tabular}{ccrrr}
    \toprule
     \bf model & \bf pretrain & \bf BLEU & \bf Pr. & \bf Rec. \\
     \midrule
     fr-top-8w & -- & 0 & 23.5 & 22.2 \\
     fr-top-10w & -- & 0 & 20.6 & 24.5 \\
     en-300h & -- & 0 & 0.2 & 5.7  \\
     fr-20h & -- & 0 & 4.1 & 3.2 \\
     \midrule
     \multirow{4}{4.5em}{\centering mb-fr-4h} & -- & 3.5 & 18.6 & 19.4 \\
      & fr-20h & 5.9 & 23.6 & 20.9 \\
      & en-300h & 5.3 & 23.5 & 22.6 \\
      & {\bf en + fr} & {\bf 7.1} & {\bf 26.7} & {\bf 23.1} \\
    \bottomrule
  \end{tabular}
  \end{center}
  \caption{Mboshi-to-French translation scores, with and without ASR pre-training. {\bf Pr.} is the precision, and {\bf Rec.} the recall score. {\bf fr-top-8w}  and {\bf fr-top-10w} are {\em naive baselines} that, respectively, predict the 8 or 10 most frequent training words. For {\bf en + fr}, we use encoder parameters from {\em en-300h} and attention+decoder parameters from {\em fr-20h}
  }
  \label{tab:bleu_mb_fr_st}
\end{table}

The baseline model trained only on ST data actually has lower precision/recall than the naive baseline, although its non-zero BLEU score indicates that it is able to correctly predict some n-grams. We see  comparable precision/recall to the naive baseline with improvements in BLEU by transferring either French ASR parameters (both encoder and decoder, {\em fr-20h}) or English ASR parameters (encoder only, {\em en-300h}).

Finally, to achieve the benefits of both the larger training 
set size for the encoder and the matching language of the decoder, we tried transferring the encoding parameters from the {\em en-300h} model and the decoding parameters from the {\em fr-20h} model. This configuration ({\em en+fr}) gives us the best evaluation scores on all metrics, and highlights the flexibility of our framework. Nevertheless, the 4-hour scenario is clearly a very challenging one.

\section{Conclusion} 
\label{sec:conclusion}

This paper introduced the idea of pre-training an end-to-end speech translation system involving a low-resource language using ASR training data from a higher-resource language. We showed that large gains are possible: for example, we achieved an improvement of 9 BLEU points for a Spanish-English ST model with 20 hours of parallel data and 300 hours of English ASR data. Moreover, the pre-trained model trains faster than the baseline, achieving higher BLEU 
in only a couple of hours, while the baseline trains for more than a day.

We also showed that these methods can be used effectively on a real low-resource language, Mboshi, with only 4 hours of parallel data. The very small size of the data set makes the task challenging, but by combining parameters from an English encoder and French decoder, we outperformed baseline models to obtain a BLEU score of 7.1 and precision/recall of about 25\%. We believe ours is the first paper to report word-level BLEU scores on this data set.

Our analysis indicated that, other things being equal, transferring both encoder and decoder parameters works better than just transferring one or the other. 
However, transferring the encoder parameters is where most of the benefit comes from. 
Pre-training using a large ASR corpus from a mismatched language will therefore probably work better than using a smaller ASR corpus that matches the output language.

Our analysis suggests several avenues for further exploration. On the speech side, it might be even more effective to use multilingual training; or to  replace 
the MFCC input features with pre-trained  multilingual features, or features that are targeted to low-resource multispeaker settings~\citep{kamper2015unsupervised,kamper+2016+arxiv+fullsegmental,thomas2012multilingual,cui2015multilingual,yuan2016learning,renshaw2015comparison}.
On the language modeling side, simply transferring decoder parameters from an ASR model did not work; but an alternative, and perhaps better, method would be to use pre-trained decoder parameters from a language model, as proposed by~\citet{ramachandran2016unsupervised+decoder+LM}, or {\em shallow fusion}~\citep{gulccehre2015using, toshniwal2018lmintegration}, which interpolates a pre-trained language model during beam search. In these methods, the decoder parameters are independent, and can therefore be used on their own. We plan to explore these strategies in future work. 


\section*{Acknowledgments}
\label{sec:thanks}
This work was supported in part by a James S McDonnell Foundation Scholar Award, a Google faculty research award, and NSF grant 1816627.  We thank Ida Szubert and Clara Vania for helpful comments on previous drafts of this paper and Antonios Anastasopoulos for tips on experimental setup.

\bibliographystyle{acl_natbib}
\bibliography{naaclhlt2019.bib}

\vfill\pagebreak

\end{document}